\documentclass[pdflatex,sn-mathphys-num]{sn-jnl}

\usepackage{graphicx}%
\usepackage{multirow}%
\usepackage{amsmath,amssymb,amsfonts}%
\usepackage{amsthm}%
\usepackage{mathrsfs}%
\usepackage[title]{appendix}%
\usepackage[dvipsnames]{xcolor}%
\usepackage{textcomp}%
\usepackage{manyfoot}%
\usepackage{booktabs}%
\usepackage{algorithm}%
\usepackage{algorithmicx}%
\usepackage{algpseudocode}%
\usepackage{listings}%
\usepackage{siunitx}

\usepackage{multicol}

\newcommand{\mb}[1]{#1}
\newcommand{\mbhat}[1]{\hat{#1}}
\renewcommand{\vec}[1]{\mathbf{#1}}

\usepackage{array,multirow,graphicx,calc}
\newlength\myheight
\newlength\mydepth
\settototalheight\myheight{Xygp}
\newcommand*\inlinegraphics[1]{%
  \settototalheight\myheight{Xygp}%
  \settodepth\mydepth{Xygp}%
  \raisebox{-\mydepth}{\includegraphics[height=\myheight]{#1}}%
}

\begin{document}

\title[Synthesizing multi-log grasp poses]{Synthesizing multi-log grasp poses in cluttered environments}


\author*[1]{\fnm{Arvid} \sur{F\" alldin}}\email{arvid.falldin@umu.se}

\author[2]{\fnm{Tommy} \sur{L\"ofstedt}}\email{tommy.lofstedt@umu.se}

\author[3]{\fnm{Tobias} \sur{Semberg}}\email{tobias.semberg@skogforsk.se}

\author[1]{\fnm{Erik} \sur{Wallin}}\email{erik.wallin@umu.se}


\author[1,2]{\fnm{Martin} \sur{Servin}}\email{martin.servin@umu.se}

\affil[1]{\orgdiv{Department of Physics}, \orgname{Umeå University}, \orgaddress{\city{Ume\aa}, \postcode{SE-90187}, \country{Sweden}}}

\affil[2]{\orgdiv{Department of Computing Science}, \orgname{Umeå University}, \orgaddress{\city{Ume\aa}, \postcode{SE-90187}, \country{Sweden}}}

\affil[3]{\orgdiv{Skogforsk (the Forestry Research Institute of Sweden)}, \orgname{Organization}, \orgaddress{\city{Uppsala}, \postcode{SE-610101}, \country{Sweden}}}

\affil[4]{\orgdiv{Algoryx Simulation}}


\abstract{Multi-object grasping is a challenging task. It is important for energy and cost-efficient operation of industrial crane manipulators, such as those used to collect tree logs from the forest floor and on forest machines. In this work, we used synthetic data from physics simulations to explore how data-driven modeling can be used to infer multi-object grasp poses from images. We showed that convolutional neural networks can be trained specifically for synthesizing multi-object grasps.
Using RGB-Depth images and instance segmentation masks as input, a U-Net model outputs grasp maps with the corresponding grapple orientation and opening width. Given an observation of a pile of logs, the model can be used to synthesize and rate the possible grasp poses and select the most suitable one, with the possibility to respect changing operational constraints such as lift capacity and reach. When tested on previously unseen data, the proposed model found successful grasp poses with an
accuracy up to 96\%.

\vspace{1em}
\textbf{Acknowledgments} \\
This work was partially supported by Mistra Digital Forest, Cranab AB, Tro\"edsson Teleoperation Lab, and the Wallenberg AI, Autonomous Systems and Software Program (WASP) funded by the Knut and Alice Wallenberg Foundation
}




\maketitle

\section{introduction}
Grasping and picking up objects comes naturally to humans but is challenging for robots. In unstructured and cluttered environments, autonomous robotic grasping requires robust perception, planning, and control. One example of this is the task of log grasping which plays an important part in modern forestry.

In cut-to-length logging, two types of machines work in pairs during the final harvest: harvesters and forwarders. Harvesters fell and cut trees into logs, and leave them distributed over the forest floor for the forwarder to collect and transport to a nearby road. The forwarding task is repetitive and exposes the operator to harmful whole-body vibrations. The machines are large and heavy, partially to protect the operator, which impacts the environment in the form of soil damage and CO$_2$ emissions.
This motivates research that targets automation and teleoperation.

\begin{figure}
    \centering
    \includegraphics[width=.7\columnwidth]{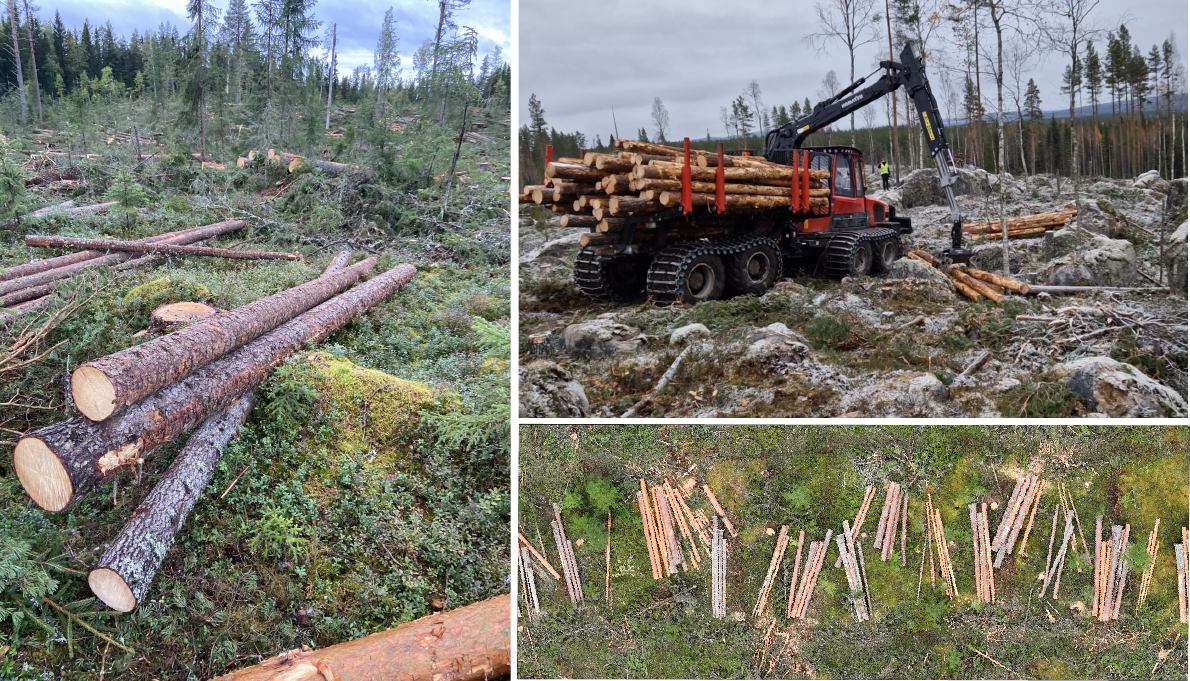}
    \caption{Left: A pile of logs jammed in between a rock and a tree stump. Top right: A forwarder grasping logs in cluttered environment. Bottom right: Log piles at a clearcut area.}
    \label{fig:photos}
\end{figure}

Autonomous forwarders would need the ability to pick up logs of arbitrary sizes from any log pile configuration, all while avoiding surrounding obstacles, see Fig.~\ref{fig:photos}. To exploit their full potential and compete with expert human performance, they must also grasp multiple logs at a time when possible. That ability requires deliberate selection of where and how to grasp in a given pile. The problem of inferring suitable grasp poses from images is known as \textit{grasp synthesis}. Since the advent of convolutional neural networks (CNNs), data-driven methods for solving it have been proposed ~\cite{morrison2020learning, kumra2020antipodal, zuo2021graph, ainetter2021end, wang2022transformer}. However, most studies consider grasping of small, everyday objects and only consider single-object grasps~\cite{cornell, jaquard}.

Multi-object grasping poses many challenges compared to the single-object case: Objects can, and must, be considered both targets and obstacles depending on one's intent. Further, objects that are within reach to be grasped together may still form an unstable or poorly balanced configuration inside the gripper, making unloading more difficult \cite{agboh2022multi}. Lastly, if grasps are inferred from a single image, then two grasps associated with the same point, $(x, y)$, may be very different in both width and orientation, and the average of valid grasps is not necessarily a valid grasp itself. This can cause problems when treating the synthesis as a regression task~\cite{chalvatzaki2021orientation}.

Log grasping can be considered a special case of multi-object grasping where the target objects may vary in size but all share roughly the same shape. Autonomous log grasping has been explored in the past few years; Andersson \textit{et al.}~\cite{andersson2021reinforcement}, Wallin \textit{et al.}~\cite{wallin2023multilog}, and Vu \textit{et al.}~\cite{vu2025autonomouswoodloggraspingforestry} used reinforcement learning to pick up logs in simulated environments, but all methods rely at least partially on access to an \textit{a priori} target pose for the crane grapple. La Hera \textit{et al.}~\cite{lahera2024exploring} and Semberg \textit{et al.}~\cite{semberg2024real} demonstrated autonomous single-log grasping on real platforms. Ayoub \textit{et al.}~\cite{ayoub2023grasp} showed how the recent progress of instance segmentation models~\cite{fortin2022instance, kirillov2023segment, steininger2025timbervision}, combined with generic grasping datasets, is applicable to form grasp plans, a method which was further explored in Ayoub \textit{et al.}~\cite{ayoub2024logloading} with impressive results on a real platform. Note that of the aforementioned studies on autonomous log grasping, only~\cite{wallin2023multilog, ayoub2023grasp, ayoub2024logloading} explored multi-log grasping and none of them considered obstacle avoidance.

We argue that models trained for single-object grasping will inevitably be limited in their ability to synthesize multi-object grasps in cluttered environments, and that models must be trained specifically for multi-object grasping to have a chance of reaching optimal performance. Furthermore, if one merely tries to identify which grasps are possible, then solutions are non-unique, and the number of solutions can easily be on the order of thousands even with a coarse discretization of the solution space. This suggests using additional metrics, other than graspability alone when searching for optimal grasps. 

In this work, we explored multi-object grasp synthesis from RGB-Depth (RGB-D) image data using physics simulations and data-driven models. Our main contributions are the following:
\begin{itemize}
    \item We show how instance segmentation masks can be used to generalize grasp synthesis to multi-object grasps in cluttered scenes.
    \item We expand the notion of grasp quality to include metrics beyond the binary quality measure conventionally used, and show how these can impact the choice of grasp during inference.
\end{itemize}

\begin{figure*}[h!]
    \centering
    \includegraphics[trim={0 0 0 0}, clip, width=\textwidth]{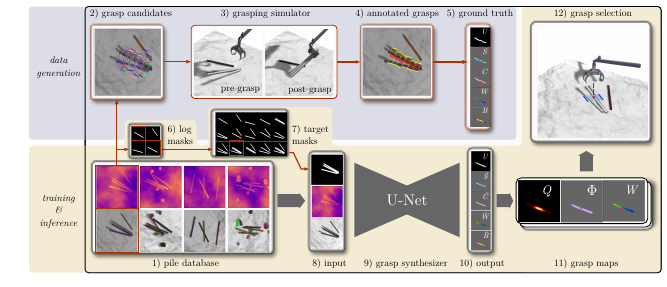}
    \caption{Model overview. 1) A sample is drawn from the pile database. 2) A rule-based algorithm is used to generate grasp candidates. 3) The grasp candidates are tested in simulation. 4) The pile is annotated with the successful grasps. 5) Grasp annotations are converted into target arrays. 6) Each log in the pile is segmented individually. 7) Individual masks are combined into target masks. With four logs, there are $2^4-1=15$ possible target subsets to consider. 8) An RGB-Depth image and a target mask are used as input. 9--10) A U-Net model is trained to predict the target variables from Step 5. Steps 2--5 and 8--10 are repeated for each of the 15 target masks. 11) The model output is used to compute the predicted grasp quality in each pixel. We search over all pixels in each of the 15 $Q$ maps and pick the grasp that maximizes the quality. 12) The chosen grasp is tested in simulation.}
    \label{fig:method}
\end{figure*}

We assumed access to an instance segmentation model such as \textit{SAM}, \textit{TimberSeg}, or \textit{TimberVision} \cite{kirillov2023segment,fortin2022instance, steininger2025timbervision}.
For each combination of target logs, a target mask was constructed from the mask of each individual target log.
Rocks, tree stumps, and non-target logs thus treated as obstacles in the cluttered scene.
To generate high-quality grasp data we developed a rule-based generator of grasp candidate poses.
The candidate grasps were evaluated in simulation and the successful ones were kept and converted into ground truth grasp maps.
The grasp maps were used to train a grasp synthesis model and its capability of predicting suitable grasps was tested on previously unseen log piles, including piles of larger size than those used for training.
The model was tested with a real crane and log piles, after which the model was further improved.

\section{METHOD}
\subsection{Method overview}
The model was trained entirely on synthetic data generated using physics simulation and 3D graphics. We began by generating a collection of artificial log pile samples and for each pile, we rendered an RGB-D image and generated instance segmentation masks for all tree logs. Via contacting multibody dynamics simulations, we were able to find examples of good grasp poses for each pile. Successful grasps were encoded into image arrays to form so-called grasp maps. We then trained a CNN model to predict grasp maps from RGB-D images. Again using simulation, we evaluated the proposed model's ability to infer high-quality multi-log grasp poses for arbitrary pile configurations. The method is summarized in Fig.~\ref{fig:method}.

Experiments were conducted with a physical forwarder loading logs in an outdoor environment. The results of the tests gave insights into shortcomings of the grasp synthesis model and ideas for improving the synthetic environment. After revising the simulator and data pre-processing, a new model was trained. It was, however, not feasible to conduct the amount of real experiments for a statistically significant test evaluation. Therefore, the test evaluation was conducted using the simulator rather than on the real machine.

\subsection{Synthesizing grasps}
Grasp synthesis is the process of producing grasp tuples, $\mathbf{\Tilde{g}} = (\mathbf{r}, \mathbf{p}, w)$, from image data, where $\mathbf{r}=(x, y, z)^\top$ is the grasp position, $\mathbf{p}\in\mathbb{R}^3$ is the gripper pose, and $w$ is the grasp width. If we assume the gripper moves only in the vertical direction, then the approach orientation can be described with a single angle, $\phi$, and the gripper's vertical position $z$ can be inferred from the depth image given a position, $(x, y)$. To be able to rank grasps, we also include a grasp quality measure, $q$, giving us the final grasp tuple definition $\mathbf{g}~=~(x, y, \phi, w, q)$.

A common approach for synthesizing grasp tuples, introduced by Morrison \textit{et al.}~\cite{morrison2020learning}, is to learn a function $\mathbf{I} \mapsto \mathbf{G}$, where $\mathbf{I}$ is an image of a scene of objects and $\mathbf{G}=(\mb\Phi, \mb W, \mb Q)$, is a \textit{grasp map}, a three-channel image of the same height and width as the input, that contains the respective grasp parameters, $(\phi, w, q)$, in each pixel. Given $\mathbf G$, an optimal grasp, $\vec g^*$, can be estimated via
\begin{align}
    (j, k) &= \underset{(m, n)}{\mathrm{arg\,max}} \: \mb Q_{mn}, \label{eq:problem}\\
    \vec g^* &= (x(j, k), y(j, k), \Phi_{jk}, W_{jk}, Q_{jk}), \notag
\end{align}
where the mapping $(j, k) \mapsto (x, y)$ usually is trivial. Grasps are assumed to be $\pi$-periodic with respect to the grasp angle $\phi$, \textit{i.e}, using $\phi=\phi_0 + n\pi$ results in the same grasp for all integers $n$.

\subsection{Input features and target maps}
As input features, we used a three-channel image $\mathbf{I} = (\mb I, \mb D, \mb{M}_T)$, \textit{i.e.}, an RGB-D image with the color channels summed and normalized to a (black and white) intensity image, together with what we call a target mask, $\mb{M}_T$.
The RGB-D image contains a top-down view of the log pile, captured with an orthographic perspective by a virtual camera placed \SI{5}{m} above the ground.
The $\mb{M}_T$ is defined as the union $\mb{M}_T = \bigcup_{i\in T}\mb{M}_i$, where $\mb{M}_i$ is a binary mask indicating all pixels that are part of the $i$th log.
The $T$ is a set that contains the indices of the logs we consider as targets.
Instead of letting the model output all valid grasps at once, we trained it to output valid grasps conditioned on that we want to pick up the logs marked by $\mb{M}_T$.
This conditioning allowed the model to associate multiple different grasps with the same pixel if needed.
Using the union $\mb{M}_T$ as input rather than individual masks $\mb{M}_i$ allows us to target an arbitrary number of logs while keeping the size of the input fixed.
Note that the model was not given access to the mask of non-target logs or obstacles, which forced it to use the RGB-D input to learn how to avoid obstructing logs, rock and stumps in the scene.

As target variables, we used a five-channel image, $G=(\mb C, \mb S, \mb W, \mb U, \mb B)$. The $\mb C$ and $\mb S$ encodes the grasp angle, $\phi$, as $C_{jk} = \cos(2\Phi_{jk})$ and $S_{jk} = \sin(2\Phi_{jk})$, respectively. This is the same component-wise encoding as the one used by Morrison \textit{et al.}~\cite{morrison2020learning}, and makes it easier for a neural network to learn the $\pi$-periodicity of the grapple orientation~\cite{hara2017designing}.
The $\mb W$ encodes the grasp width $w$ in each pixel and ranges from \SI{0.30}{m} to \SI{1.55}{m}.
The $\mb{U}$ is an indicator image, with pixels close to a successful grasp labeled 1 and others 0. While perfectly binary in the ground truth data, it can take on any value in $(0, 1)$ in the model output. To emphasize the difference, we will refer to this value as \textit{graspability} when talking about the model's continuous output.

\begin{figure}
    \centering
    \includegraphics[trim={0 .5cm .5cm 0}, clip, width=0.6\columnwidth]{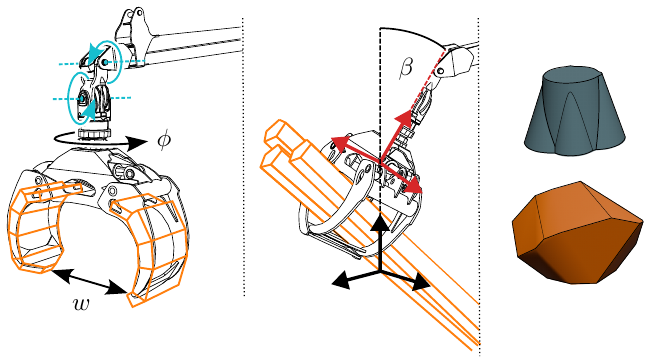}
    \caption{Definition of the grapple width, $w$, grapple orientation, $\phi$, and balance angle, $\beta$. The simplified collision geometry of the grapple and tree logs are drawn with orange lines. Cyan arcs show the rotators's unactuated joints. Examples of obstacle geometries are shown on the right.}
    \label{fig:balance}
\end{figure}

To be able to separate good grasps from the ones that are merely possible, we introduce a \textit{balance map}, $\mb{B}$. The balance map encodes the grapple balance value, which is defined as the cosine of the angle $\beta$ that the grapple makes with the world $z$-axis after a completed grasp.
Definitions of the grapple angle, width, and balance are illustrated in Fig.~\ref{fig:balance}.

The indicator image $\mb{U}$ is what is commonly denoted as the quality map $\mb Q$ in grasp synthesis. However, in multi-object scenes, the number of valid grasps is typically large, and we argue that a binary metric is insufficient when forming more sophisticated grasp plans. Therefore, we extended the notion of grasp quality to include other metrics via an objective function $f$ acting element-wise:
\begin{equation}
    \mb Q_{jk} = f(\mb U_{jk}, \tau, \mb B_{jk}), \label{eq:grasp_quality} 
\end{equation}
where $\tau$ is the number of target logs --- which is the same as the number of logs included in the target mask. Recall that during inference, we chose the grasp that maximizes $\mb{Q}_{ij}$ in \eqref{eq:problem}. Hence, we could change our definition of grasp optimality by considering different choices of $f$. And, since the neural network outputs $\mb{U}$ and $\mb{B}$, rather than $\mb{Q}$ directly, we were free to change our definition of $f$ without having to retrain the neural network.

We explored three different choices of objective functions, $f$:
\begin{align}
    f_1(u) &= u \\
    f_2(u, \tau; \mu) &= u\tau^\mu \\
    f_3(u, b, \tau; b_\text{opt}, \sigma_b) &= u\tau^\mu\exp(-(b - b_\text{opt})^2/\sigma_b^2).
\end{align}
The $f_1$ promotes graspability only. This is equivalent to the conventional definition of grasp quality. The $f_2$ aims to maximize the expected number of grasped logs, and we found empirically that using $\mu=\tfrac{1}{3}$ gave a good trade-off between graspability and the number of target logs. Finally, $f_3$ aims to promote well-balanced grasps with high graspability by weighting the quality by how close the predicted balance value is to optimal balance. The $b_\text{opt}$ denotes which value we consider to be optimal, and $\sigma_b$ how harshly we penalize grasps that deviate from $b_\text{opt}$. We chose to use $b_\text{opt}=1$ as the optimal balance value and found empirically that $\sigma_b=0.25$ was a suitable value.

\subsection{Model architecture and training}
We estimated the function $\mathbf{I} \mapsto \mathbf{G}$ using a 4.1M parameter UNet~\cite{ronneberger2015unet}. As loss, $\mathcal{L}$, we used a linear combination of binary cross entropy, $\mathcal{L}_\textsc{BCE}$, and what we refer to as a \textit{masked mean squared error} (MMSE), $\mathcal{L}_\textsc{MMSE}$:

\begin{align*}
    \mathcal{L}&(\mb C, \mbhat C, \mb S, \mbhat S, \mb W, \mbhat W, \mb U, \mbhat U, \mb B, \mbhat B)= \\
    &\mathcal{L}_\textsc{BCE}(\mb U, \mbhat U) +\lambda_\mb C\mathcal{L}_\textsc{MMSE}(\mb U, \mb C, \mbhat C)+\lambda_\mb S\mathcal{L}_\textsc{MMSE}(\mb U, \mb S, \mbhat S) \\
    &+\lambda_\mb W\mathcal{L}_\textsc{MMSE}(\mb U, \mb W, \mbhat W) +\lambda_\mb B\mathcal{L}_\textsc{MMSE}(\mb U, \mb B, \mbhat B),
\end{align*}
where the $\lambda$s are regularization parameters, and hatted variables denote model outputs. The MMSE loss function is defined as
\begin{equation}
    \mathcal{L}_\textsc{MMSE}(\mb{U}, \mb{Y}, \mb{\hat{Y}}) = \frac{1}{N^2}\sum_{j=1}^N\sum_{k=1}^N u_{jk}(y_{jk} - \hat{y}_{jk})^2,
    \label{eq:loss_mmse}
\end{equation}
where $N$ denotes the image dimensions (both height and width). By taking the elementwise product with the ground-truth indicator map, $\mb U$, in \eqref{eq:loss_mmse}, we only penalize predictions made in pixels that we know are valid grasp points. Hence, we did not have to make any assumptions about which value to assign pixels outside valid grasps, which would have been a concern with any regular, unmasked, loss function. Note that the neural network will output values for all pixels, but values outside the grasp zone are ignored by the loss. Our best model was trained using $[\lambda_\textsc{B}, \lambda_\mb S, \lambda_\mb C, \lambda_\mb W] = [120, 30, 30, 60]$, but model performance was found to be robust with respect to small changes to the regularization parameters.

We trained on 80\% of 5,491 annotated samples and used the remaining 20\% as the validation set. Testing was done on a separate set of 1,200 unannotated piles. During training, we used random flips and rotations, randomized colors and lighting, and Gaussian noise to augment the data.

\subsection{Grasp candidate reduction}
It is a challenge in itself to automatically generate consistent training data when the solutions are non-unique --- even in a simulated environment. Our approach was to first run a rule-based search for promising grasps to create a list of what we refer to as \textit{grasp candidates}. We then tested those candidates in simulation until we found enough successful grasps.

The initial search for grasp candidates was entirely based on geometry and did not take any dynamics into account. We emphasize that this search algorithm, while helpful during data generation, would not work for grasp synthesis in the wild as it relies on having access to the exact position, pose, and shape of each tree log and obstacle. Moreover, even with access to perfect information, we were not able to formulate a simple set of rules that guaranteed successful grasps. This led us to simulate thousands of failed grasp trials during data generation.

The number of grasp candidates for a given target subset of logs could range anywhere from zero to a couple of thousands. When the number of candidates was large, the candidates included many grasps that were almost identical, as well as grasps that were needlessly wide or awkwardly oriented. To reduce the candidate list to a few high-quality grasps, we used Algorithm~\ref{alg:grasp_alg} in the simulations.

\begin{algorithm}
    \caption{Grasp candidates reduction}\label{alg:grasp_alg}
    \begin{algorithmic}[1]
        \Require A set, $G$, of candidate grasp tuples
        \Ensure A set, $S$, of successful grasps
        \State $S \gets \varnothing$
        \While{G is not empty}
            \State Select the narrowest grasp, $g$ from $G$
            \State Simulate grasp $g = (x, y, \phi, w)$
            \If{$g$ successful}
                \State Sample $\Delta w \sim \mathcal{U}(0.05, 0.10)$ (cm)
                \State $g' \gets (x, y, \phi, w + \Delta w)$
                \State Simulate $g'$
                    \If{$g'$ successful}
                        \State $g \gets g'$
                    \EndIf
                \State $S \gets S \cup \{g\}$ \Comment{add $g$ to $S$}
                \State Remove grasps from $G$ that overlap with $g$.
            \EndIf
            \State $G \gets G \setminus \{g\}$ \Comment{remove $g$ from $G$}
        \EndWhile
    \end{algorithmic}
\end{algorithm}
In step 7 of Algorithm~\ref{alg:grasp_alg}, we compute the area of the grasp rectangle intersection between $g$ and all remaining candidates. If the overlap exceeds a threshold value (\SI{0.04}{m^2}), the candidate grasp is discarded. 
Algorithm~\ref{alg:grasp_alg} typically terminated after finding a handful of non-overlapping grasps. Giving priority to the narrowest grasp candidates has two advantages: it naturally places the grasp in the middle of the two outermost target logs, and helps pick out the grasp that is best aligned with the targeted logs. On the other hand, choosing the narrowest possible grasp leaves little room for error in positioning (see section ~\ref{sec:experiments}) and we add lines 6--11 to select wider grasps if the situation allows for it.


\subsection{Simulator}
\begin{figure*}
    \centering
    \includegraphics[width=0.24\textwidth]{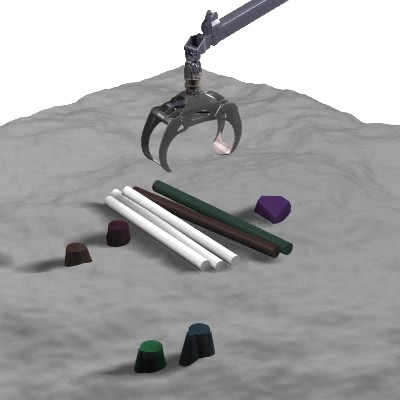}
    \includegraphics[width=0.24\textwidth]{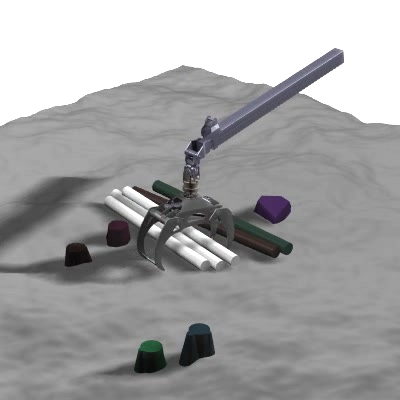}
    \includegraphics[width=0.24\textwidth]{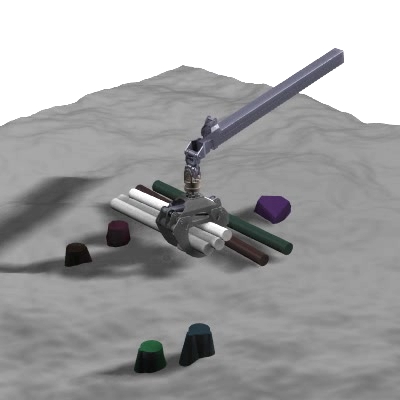}
    \includegraphics[width=0.24\textwidth]{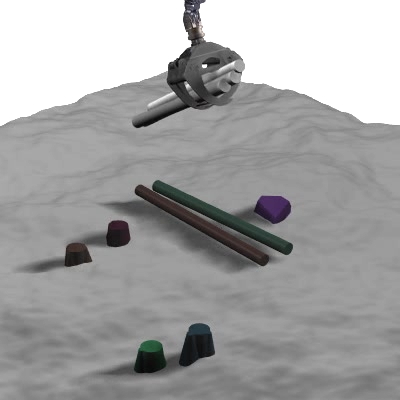}
    \caption{Example of a simulated grasp. The grapple is lowered over the target logs (white)  using a grasp pose (position, orientation, opening width) that avoids collision with the surrounding logs, stumps and rocks (colored).}
    \label{fig:simulator}
\end{figure*}

We generated synthetic data and evaluated the model using multibody dynamics simulations run using the physics engine AGX Dynamics~\cite{AGX2021}.
The simulated environment consisted of a terrain, tree logs, tree stumps, rocks, and a crane grapple. An example of a simulated grasp is shown in Fig.\ref{fig:simulator} .The terrain was represented by a \SI{5}{m}$\times$\SI{5}{m} heightfield, producedurally generated using Perlin noise~\cite{perlin85image}.
Logs were modeled as having a cylindrical rendered geometry but a pentagonal prism as collision geometry, see Fig.~\ref{fig:balance}.
The pentagonal cross-section provided rolling resistance, which made the log-log interactions less smooth, and presumably less predictable. The logs' lengths, $\ell \sim \mathcal{N(\SI{2.5}{m}, \SI{0.2}{m})}$, and diameters, $d \sim \mathcal{N(\SI{0.16}{m}, \SI{0.02}{m})}$, were sampled from normal distributions.
Tree stumps were modeled using a large truncated cone as the main body, and smaller truncated cone placed randomly along the main body's perimeter to emulate roots and giving the stumps a more irregular shape. The height and radius of the stumps were sampled from normal distributions.
Rocks were modeled as irregular polyhedra with randomized orientation and size, with diameters ranging from \SI{0.3}{m} to \SI{1.3}{m}.

We modeled the grapple and rotator as a system of nine rigid bodies held together by ten joints.
The rotator was attached to a boom which was under kinematic control.
The rotator connected the boom and the grapple via three hinge joints, of which only one was actuated.
The two unactuated hinge joints made the grapple compliant, meaning it could for example `fold' if pressed too hard against a log or obstacle, or bounce off after high-impact collisions.
The grapple's design was based on a scaled-down version of a Cranab CR HD grapple~\cite{Cranab2023}, and we tuned its closing strength to roughly match that of its real counterpart.
As with the logs, we replaced the grapple's collision geometry with a collection of cuboids to make contact computations faster and more stable.
The system is illustrated in Fig.~\ref{fig:balance}.

Pile samples were generated by first spawning a piece of terrain, placing 0--7 obstacle at random positions, and then dropping a stack of logs in the scene and letting them settle into a stable configuration. The data generation process was the same for all piles except for some of the case study piles.

Logs, terrain and obstacles were individually given random colors of intensities between 0.2 and 0.8. As a form of data augmentation, three versions of every scene was rendered, each with its own randomized colors and light conditions.


Once grasp candidates had been generated for each subset of target logs, we performed grasp trials to confirm if they were indeed viable grasps.
A trial started with the grapple positioned directly above the grasp point.
We lowered the grapple until the number of logs inside of the grapple was equal to the number of target logs, closed the grapple as much as the force limits allowed us to, and raised the grapple until it had returned to its starting height.
A trial was classified as failed if the set of logs ending up in the grapple was not the target logs, if the grapple was not sufficiently closed, or if a collision force acting on the grapple at some point exceeded a threshold value.
The second condition was enforced as it is generally advised against lifting logs if the grapple claws do not enclose the load completely.
The last criterion helps filter out grasps that would be considered reckless and could potentially damage the machine in a real scenario.

Any grasp data we generated would inevitably be biased towards the controller that was used to create them, but we argue that this bias is reduced by using as simple a controller as possible.
To further decrease the control dependency, we disabled all contacts between the terrain and grapple.

\subsection{Synthetic data set}
We simulated 604 piles, each consisting of four logs. The decision to only use piles of four logs during training was deliberate, as it allows us to test the model's ability to generalize to both more and fewer target objects than it was trained on.

A third of the piles featured one or more obstacle. With four logs, there are $2^4-1=15$ possible target subsets to consider, and we tried to find valid grasps for each subset.
After removing samples that displayed unphysical behavior or unstable log configurations, we were left with 5,491 grasp map samples, annotated with 12,124 grasps in total. Out of the 5,491 samples, 2,039 were cases where no successful grasp could be found for the given targets. In addition to these 604 piles, we generated a test set of 1,200 unannotated piles.

\begin{figure}
    \centering
\includegraphics[width=0.5\columnwidth]{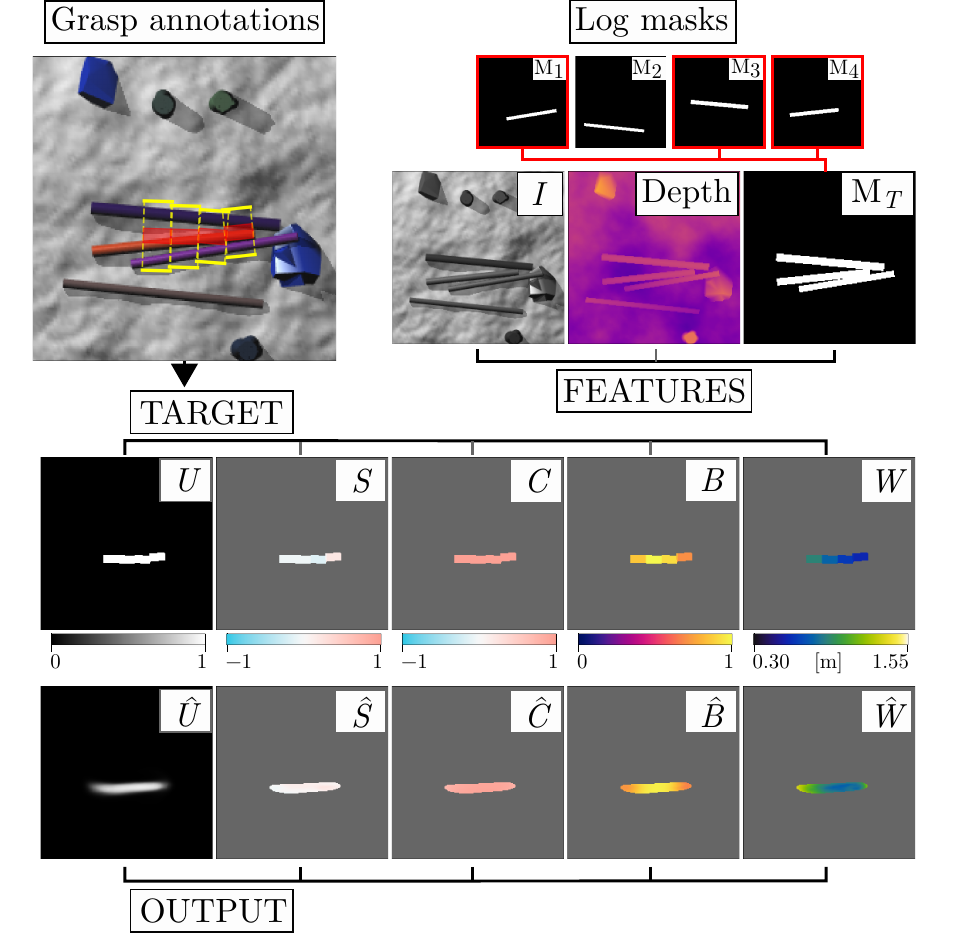}
    \caption{Illustration of how annotated grasps are encoded into 2D arrays. \textbf{Top left:} The thick yellow lines show the position of the grapple claw tips, and the red areas show the rectangle that we used to encode the grasp parameters in the target arrays. The bottom row shows an example output from the neural network.} 
    \label{fig:encoded_sample}
\end{figure}

Each sample consisted of an RGB-D image and a segmented mask of target logs, along with grasp data encoded into 2D arrays as seen in Fig.~\ref{fig:encoded_sample}. The encoding was made using the grasp rectangle method proposed by Morrison \textit{et al.}~\cite{morrison2020learning}. Unlike them, we used rectangles of constant width (\SI{20}{cm}) rather than letting the rectangles' sizes be proportional to the grasp width, $w$.

\subsection{Experiments on a real forwarder}
\label{sec:experiments}
Experiments with a real forwarder crane were carried out during the development phase using an early version of the grasp synthesis model, denoted V0. The V0 model did not have access to depth data as input and was trained on a slightly different dataset (see below) but was otherwise identical in structure to the final version models presented in the Results section.

The experiments were conducted using the Extractor XT28 concept forwarder at Troëdsson teleoperation laboratory outside Uppsala. The forwarder is equipped with a crane consisting of a Loglift 91F boom, a double swing damper
(Indexator Dual swing damper 80–80-45 HD) and 0.35 m2-grapple from Loglift (FX 36)~\cite{semberg2024real}. The system includes also a GNSS receiver (ArduSimple, ublox F9p) with RTK-correction from the Swedish Land Survey Authority
(Lantmäteriet) and a single stereo camera (Stereolabs ZED 2). 

\begin{figure}
    \centering
    \includegraphics[width=0.49\linewidth]{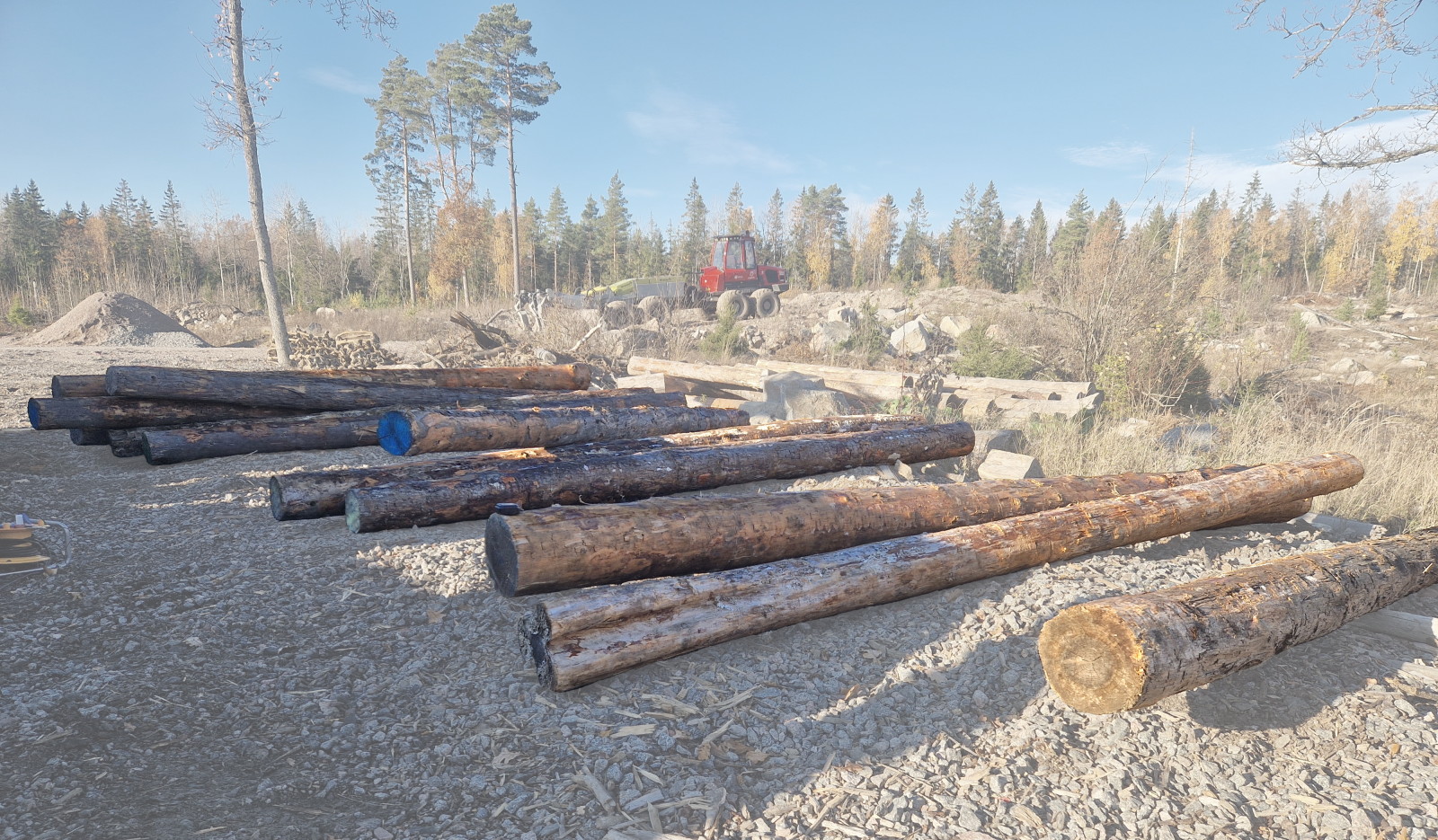}\vspace{1mm}
    \includegraphics[width=0.49\linewidth]{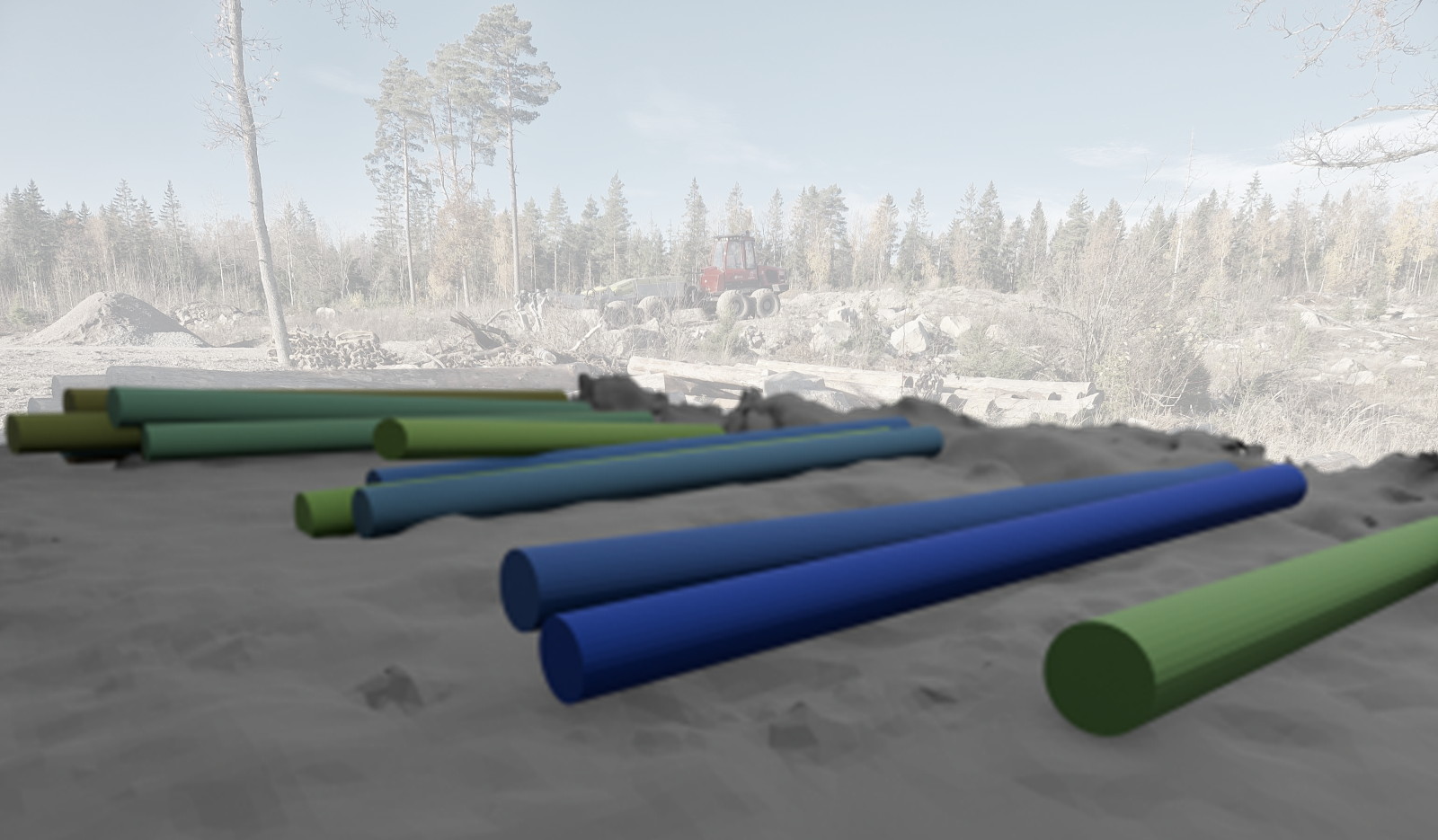}
    \caption{Sample set of logs piles and the corresponding 3D reconstruction used for creating the RGB inputs for the grasp synthesizer.}
    \label{fig:scene_reconstruction}
\end{figure}

\begin{figure}
    \centering
    \includegraphics[width=\linewidth]{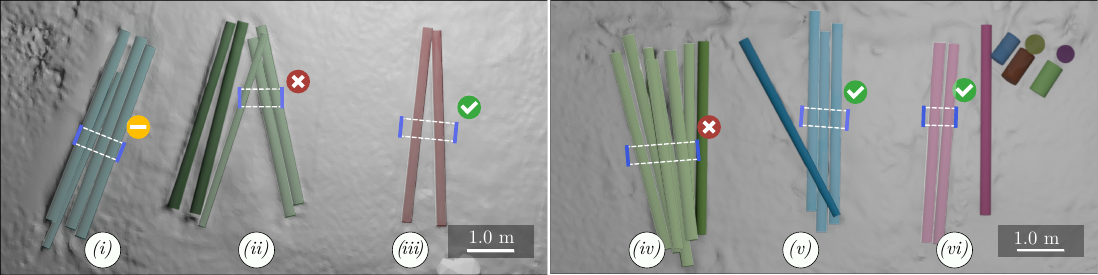}
    \caption{The six piles used for the real experiments. Green markers denote successful grasps, red marker failed (due to collision) and yellow markers denote successful but suboptimal grasps.}
    \label{fig:experiments}
\end{figure}

Six log piles were created on a flat gravel terrain by manually placing the logs in configurations typical to a harvesting site. The number of logs in each pile ranged between two and seven. The person placing the logs was not involved in the model development and hence, had no knowledge about the V0 model's strengths and weaknesses. The piles were scanned using the stereo camera, and a number of log endpoints were 3D positioned using the GNSS receiver. The scenes were reconstructed in the simulation environment, see Fig.~\ref{fig:scene_reconstruction} and orthographic top-down images on the same format as the ones the models were trained on were rendered, see Fig.~\ref{fig:experiments}. The V0 model was evaluated on the synthetic images, and the best grasp suggested by the model for each pile was transformed from the local image coordinates into GPS coordinates. During experiments, the crane was controlled manually, starting with the gripper placed roughly one meter above the logs in the position and pose suggested by the model. The operator was instructed to lower the gripper in a purely vertical motion until one of the tips made contact with the ground, close the grapple, and then lift the logs one meter up in the air. 

\begin{figure}
    \centering
    \includegraphics[width=0.49\linewidth]{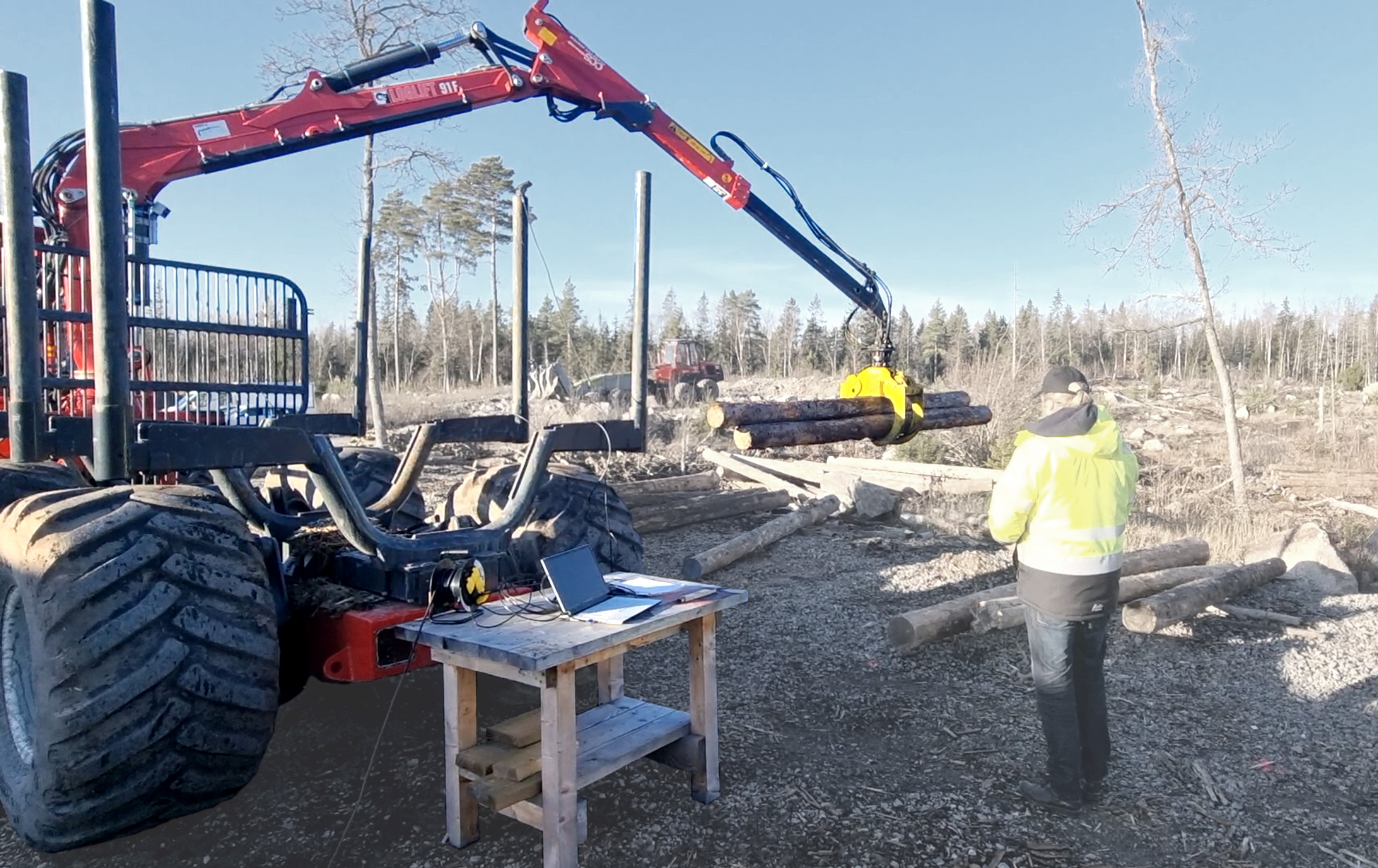}\vspace{1mm}
    \includegraphics[width=0.49\linewidth]{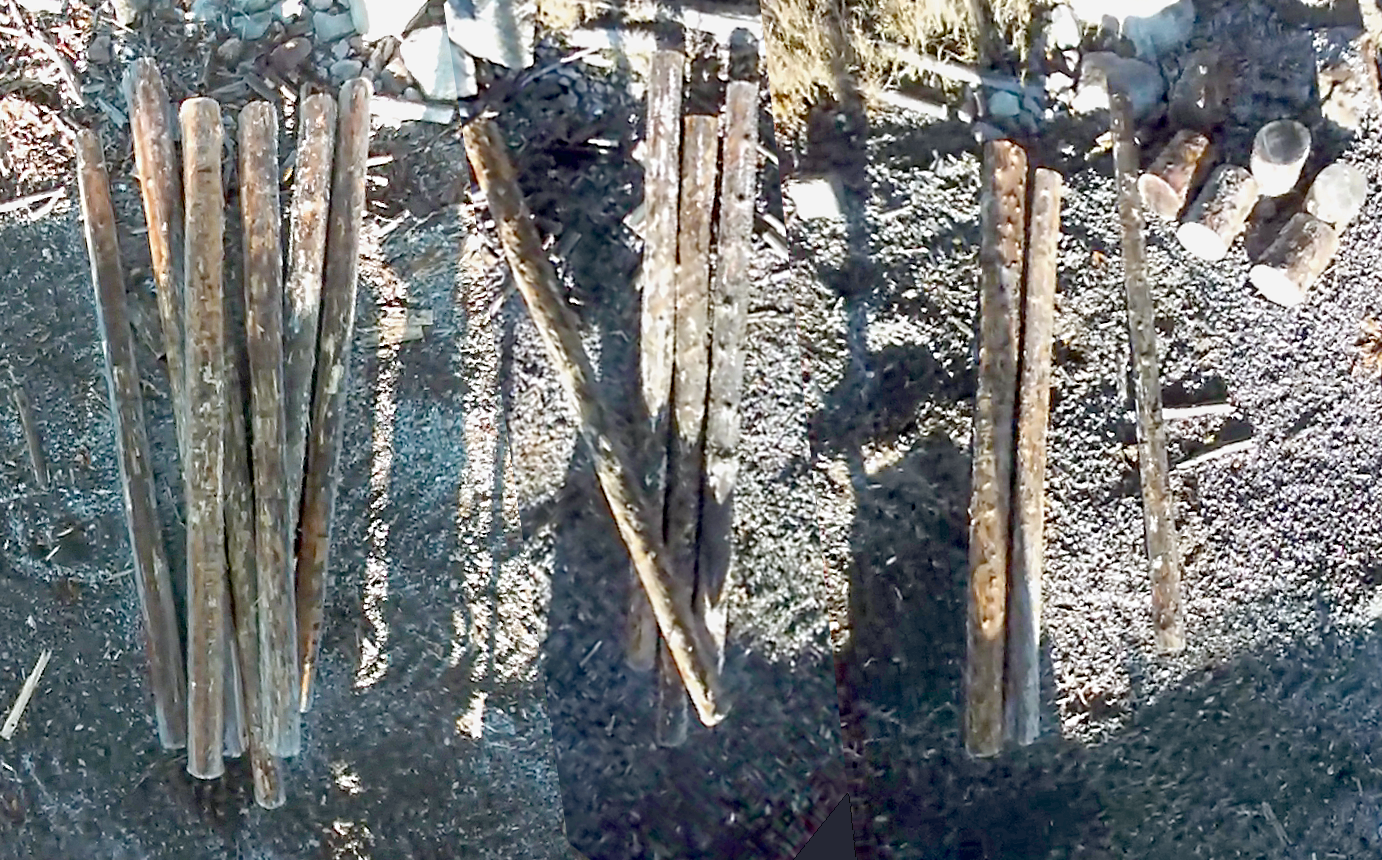}
    \caption{The experiment setup, including the forwarder crane (left) and th logs piles (right).}
    \label{fig:exp_setup}
\end{figure}

Out of six trials, four resulted in successful grasps, see Fig.~\ref{fig:experiments}. \textit{(ii)} failed due to a combination of a small error in positioning along with a too narrow grasp width, which resulted in grasping only two logs of the indented three.
The grasp in pile \textit{(iv)} in Fig.~\ref{fig:experiments} is also classified as a failure because one of the logs fell out of the grapple during lifting, resulting in five grasped logs instead of the intended six. 
In one of the successful trials (\textit{(i)} in Fig.~\ref{fig:experiments}), the proposed grasp was too narrow to encircle all targets, and the trial only succeeded because the weight of the grapple made the logs roll into it.

The V0 model exhibited a bias towards 
narrow grasps in both simulation and experiments. To mitigate the bias, we made adjustments to the Grasp candidate reduction algorithm (Algorithm \ref{alg:grasp_alg}), adding lines 6--11 to encourage wider grasps when the situation allows it. Using a grasp wider than necessary makes grasping more robust to error in positioning, but increases the risk of colliding with adjacent objects. The range for the added grasp width in Algorithm~\ref{alg:grasp_alg} (\SI{5}{cm}--\SI{10}{cm}) roughly corresponds to the precision in the GPS positioning.

\section{RESULTS}
The results section is structured as follows: First we present the model's overall performance on the test set when we select grasps using the conventional metric for grasp quality. Secondly, we show that the performance can be improved by using other objective functions during inference. Thirdly, we take a closer look at how the model works on some selected special cases. And finally, we test our model's generalization ability via a number of perturbation tests. 
In our figures we emphasize the difference between output and outcome. The \textit{output} is the top grasp suggested by the model and these are displayed in the figures using grasp rectangles. The \textit{outcome} is the result of testing the suggested grasp in simulation, and can be classified as either successful, failed, or inconclusive. We use the symbols \inlinegraphics{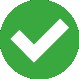}, \inlinegraphics{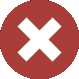}, \inlinegraphics{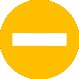} to indicate these in our figures, respectively. Inconclusive grasps are typically ones that succeed more because of a fluke (dynamics) rather than because the grasp pose was particularly good. In all sections and figures we use the objective function $f_1$ to select the optimal grasp unless otherwise stated.

\subsection{Overall performance}
We tested the proposed model in simulation on a set of 1,200 unannotated log piles of varying size, with half of the test samples featuring obstacles. The performance is summarized in Table~\ref{tab:results_obj} and examples of both failed and successful grasps are found in the Appendix. The model generalizes well to piles of sizes equal or smaller than the ones it was trained on, and fairly well to larger piles, however the fail rate increases with the number of logs. Note that the success rate increases when the number of logs is reduced, suggesting that part of the drop in success rate for larger piles is due to an increase in task difficulty rather than generalization error only. Note also that the average number of grasped logs increased with pile size despite the increased fail rate.

Looking closer at the 125 samples where the model failed we found that the most common mode of failure was grasping more logs than intended (84) followed by colliding with obstacles or non-target logs (34). The remaining seven failures were due poor grasp poses (too narrow, bad orientation) and/or logs falling out of the grapple during the lifting phase.

\begin{table}[h]
    \centering
     \caption{Test set performance (success rate/average number of grasped logs) for the two considered models for different pile types. For each combination of pile size and obstacles (Yes/No), we simulate 100 piles, one grasp per pile, giving us 1,200 piles in total.}
    \begin{tabular}{m{1.3cm}m{1.4cm}m{1.1cm}m{1.1cm}m{1.1cm}m{1.1cm}m{1.1cm}m{1.1cm}}
    \toprule Obstacles & 2 logs & 3 logs & 4 logs & 5 logs & 6 logs & 7 logs & Total \\
    \midrule
    Yes & 99\%/1.7 & 87\%/1.9 & 89\%/2.7 & 89\%/3.6 & 83\%/3.9 & 78\%/4.5 & \multirow{2}{*}{90\%/3.1} \\
    \cmidrule(lr){1-7}
    No & 100\%/1.8 & 99\%/2.5 & 98\%/3.2 & 90\%/3.6 & 89\%/4.1 & 73\%/4.2 & \\
    \bottomrule
    \end{tabular}
    \label{tab:results_2_to_7}
\end{table}

\subsection{Objective functions}
We test using different objective functions to select our grasps on the same 1,200 piles as presented in Table \ref{tab:results_2_to_7}. The results are summarized in Table~\ref{tab:results_obj}. We find that using $f_2$ or $f_3$ indeed does improve the average number of grasped logs, and that $f_3$ also improves the average balance of grasps. Note that these gains does come at the expense of success rate, on the contrary, the success rate is improved.

\begin{table}[]
    \centering
     \caption{Test set performance in simulation. $\beta_\text{avg}$ is the average balance angle. Balance statistics are calculated using successful grasps only.}
    \begin{tabular}{m{1cm}m{1cm}m{1.5cm}m{.6cm}}
    \toprule Quality \newline function & Success \newline rate & Avg. no. of \newline grasped logs & $\beta_\text{avg}$ \\
    \midrule
    $f_1$ & 90\% & 3.1 & \ang{12} \\
    \midrule
    $f_2$ & 95\% & 4.0 & \ang{11} \\
    \midrule
     $f_3$ & 96\% & 4.0 & \ang{6} \\
     \bottomrule
     \end{tabular}
     \label{tab:results_obj}
\end{table}



\subsection{Case study --- Regular pile}
\begin{figure*}
    \centering
    \includegraphics[width=\textwidth]{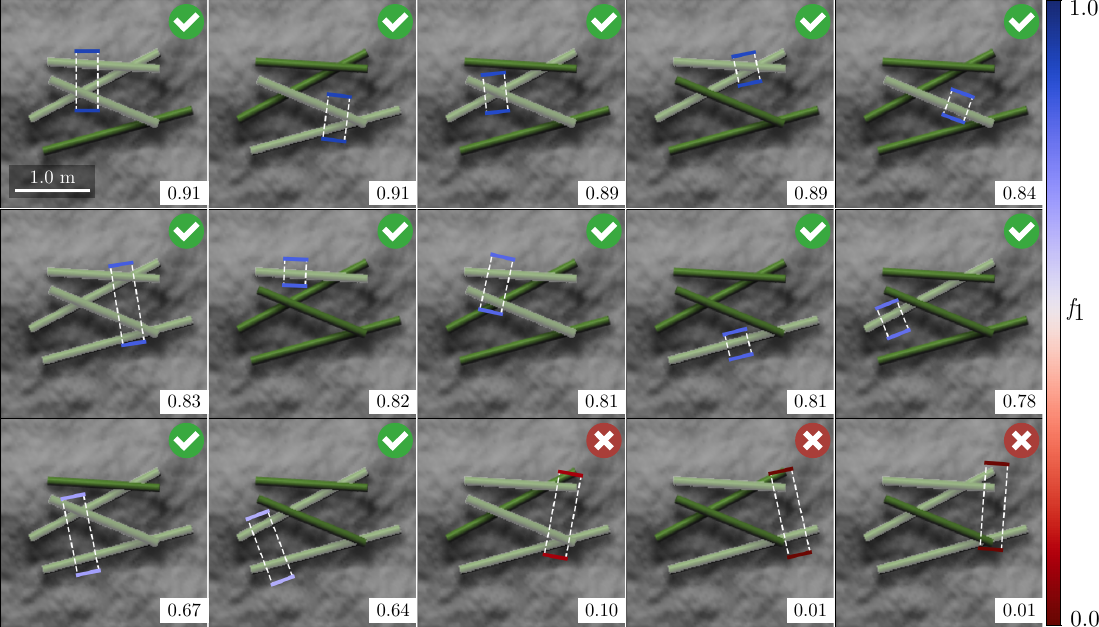}
    \caption{Optimal grasp for each possible target subset in a pile. Target logs have been colored white for illustrative purposes. The subsets are ranked from best to worst, and the grasps are color-coded by their graspability score, $f_1$, which is also displayed in the bottom right of each subfigure.}
    \label{fig:all_subsets}
\end{figure*}

To understand how the proposed model is meant to be used, consider the single pile shown in Fig.~\ref{fig:all_subsets}. In each subfigure, we used a unique set of target logs as input to the model. We made one evaluation for each target subset and found the best grasp for each one (here using the objective function $f_1$). We used the model's predicted quality score to rank the different subsets, and as seen in Fig.~\ref{fig:all_subsets}, the best grasp suggested by the model is targeting the three uppermost logs.

\subsection{Case study --- Obstacles}
\begin{figure}
    \centering
    \includegraphics[width=0.5\columnwidth]{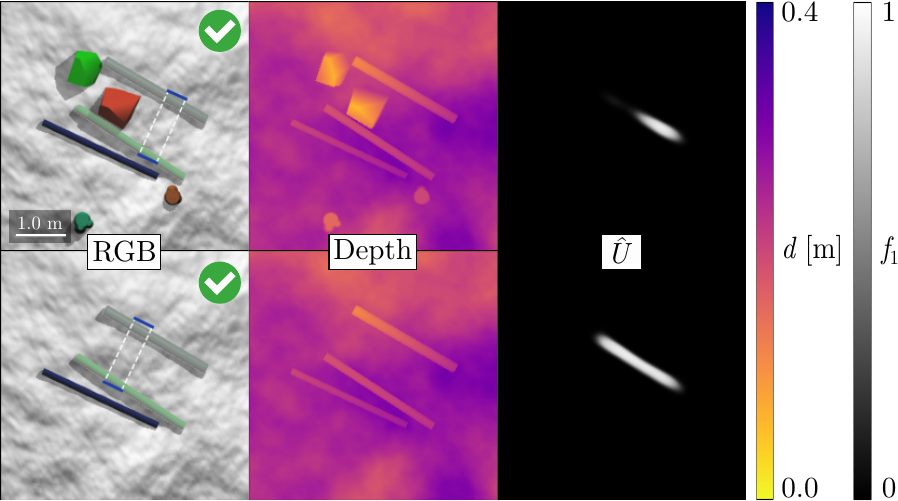}
    \caption{Example of how the model responds to obstacles in the scene. RGB and depth images are rendered twice for the same scene, first with obstacles and then without, and the model is evaluated on both versions for comparison. The right-most column shows the model's indicator output for each version. Note how the model avoids grasping close to the large boulder in the top row.}
    \label{fig:obstacle_impact}
\end{figure}

To probe how the model responds to obstacles we created pile samples in two versions, one with obstacles and one without. We rendered RGB and Depth data for the two versions, evaluated the model on them both, and observed the difference in output. One such comparison is shown in Fig.~\ref{fig:obstacle_impact}, where it is apparent that the model has learned to associate large rocks with a lower grasp probability. However, removing obstacles did not always have as large impact as in Fig.~\ref{fig:obstacle_impact}. The impact was typically larger for larger obstacles, and the difference in output could be close to zero for cases with smaller obstacles. When looking closer at samples with failed outcomes due to obstacle collisions we find that these were also cases where the model produced more or less identical output when the pile was rerendered without any obstacles.

\subsection{Case study --- Packed logs}
\begin{figure}
    \centering
    \includegraphics[width=0.5\columnwidth]{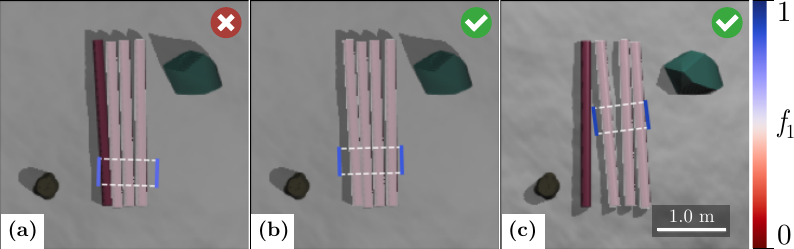}
    \caption{Model performance on tightly packed piles. Targeted logs are indicated by a brighter color. The model performs poorly when tasked with targeting a subset of logs in a tightly packed pile (a). The usual (high) performance is recovered if all logs are targeted (b) or if there are small gaps in between some of the logs (c). This rarely causes problems, as the model consistently assigns higher $f_1$ score to grasps targeting all logs.}
    \label{fig:packed}
\end{figure}

In practice, it is common to find logs placed in aligned and tightly packed piles and it was therefore of interest to see how the proposed model performs in such scenarios. A small set of tightly packed piles of four logs was created and evaluated using our model. When the model was allowed to to choose targets freely, it would always choose to grasp all four logs at once, see Fig.\ref{fig:packed}a. However, when constraining the model to only target a subset of the logs, it would consistently propose too wide grasps, resulting in either grasping more logs than intended, or, the grapple colliding with adjacent logs, see Fig. \ref{fig:packed}b. The inability to grasp subsets usually disappeared as soon as small gaps were introduced between the logs, see Fig.\ref{fig:packed}c.


\subsection{Perturbation tests \& Ablation study}
\begin{figure}
    \centering
    \includegraphics[width=0.5\columnwidth]{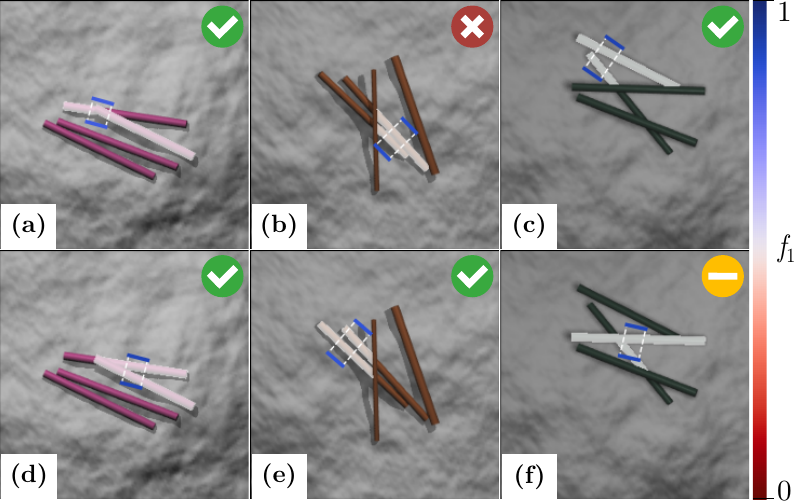}
    \caption{Example output from the mask perturbation tests. \textbf{(a)--(e)}: Examples of feasible grasps suggested by the model given incomplete target masks (shown in white). \textbf{(f)} Example of a questionable grasp assigned a high $f_1$ score by the model. \textbf{Comment: consider only showing top row. Reduce font size}}
    \label{fig:incomplete_masks}
\end{figure}

As the proposed method assumes access to segmented masks of each target log, it was motivated to investigate how the model would respond to errors in the target mask. More specifically, we investigate the effect of feeding the model incorrectly split-up masks caused by occluding objects. We found the model to be robust to the introduced errors, and it typically output grasps such as the ones shown in Fig.~\ref{fig:incomplete_masks}a--e, whereas erroneous output, such as the colliding grasp shown in Fig.~\ref{fig:incomplete_masks}f, were rare. In these tests, all logs were given the exact same color, as the color could otherwise indicate which pieces are connected to each other.

\begin{figure}
    \centering
    \includegraphics[width=0.5\columnwidth]{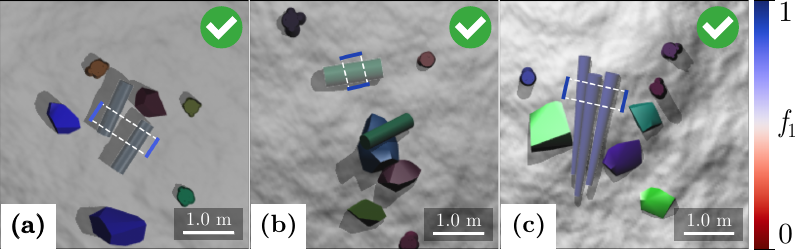}
    \caption{Example output from the size and shape perturbation tests. \textbf{(a)--(b)}: Best overall grasp according to the model when evaluating on short and thick tree logs. \textbf{(c)}: Best grasp suggested by the model for a pile of conical logs.}
    \label{fig:pert_size_shape}
\end{figure}

We also tested how well the model generalizes to other shapes and size of the tree logs. Fig.~\ref{fig:pert_size_shape}a \& ~\ref{fig:pert_size_shape}b show model output when applied to logs much thicker and shorter than the logs it was trained on ($>$ six standard deviations from the training set means). Fig.~\ref{fig:pert_size_shape}c shows the model's top suggestions for all target subsets in a pile of conical logs. From these tests we see that the model is able to generalize beyond the cylindrical logs it was trained on.

We also trained a version of the model that did not use depth as input. The performance was only slightly worse for the model that did not have access to the depth information. A summary of these results can be found in the Appendix.


\section{Transferability to a real system}
The model is trained on synthetic images and will not by itself generalize to real images. In principle, one could annotate real pile images with grasp rectangles and add them to the dataset, either for fine-tuning or for retraining as part domain randomization framework. However, real images feature much higher variation than our synthetic ones and it would likely require many more real samples to reach comparable performance. Perhaps a more appealing solution is to automate the 3D reconstruction process described in section \ref{sec:experiments}, similar to Ayoub \textit{et al.}~\cite{ayoub2023grasp}. Another is to use generative models to either augment the images during training~\cite{chen2023genaug} or make real images look like synthetic ones during inference~\cite{rao2020rlcyclegan}. 

\section{Conclusions}
We conclude that generative grasp models can be trained specifically for multi-object grasping, and that instance segmentation masks can be used to help a model distinguish between targets and obstacles in cluttered scenes. Additionally, the use of segmentation masks allows us to act with intent, making the model output easier to interpret. The use of a single-channel target mask makes the model formulation general with respect to the number of target objects, and we have shown that the model is able to detect grasp poses for more targets than it was ever exposed to during training. 


While the proposed method does not replace the need for intelligent control, we believe it has the potential to boost the performance of control solutions that rely on access to grasp poses, such as the one proposed by Wallin et al.~\cite{wallin2023multilog}. 

The model struggles to produce good grasps for picking up subsets of logs from tightly packed piles. The performance could potentially be improved by adding more training data specifically targeting these cases but we suspect that the deficiency is at least partly due to the simplistic controller used in this study, and that grasping from large stacks of packed logs probably requires more adaptive control. Apart from this, the main limitation of the proposed method is perhaps that the number of possible target subsets increases exponentially with the total number of logs in a pile. Evaluating all possible target subsets would inevitably be computationally limiting as piles grow large. In cases where an exhaustive search over all subsets is not feasible, one could imagine having some pre-processing step where only the most relevant subsets are selected for inference, \textit{e.g.}, based on proximity to the manipulator's base. 

\section{Code \& data availability}
Data sets and models generated during the current study are available from the corresponding author on reasonable request.



\bibliography{IEEEabrv, ref}

\section{Statements \& Declarations}
\subsection{Competing interests}
The authors declare that they have no known competing financial interests or personal relationships that could have appeared
to influence the work reported in this paper.

\subsection{Funding}
This work was partially funded by Mistra Digital Forest Grant DIA 2017/14 \#6. Cranab AB, Tro\"edsson Teleoperation Lab, and the Wallenberg AI, Autonomous Systems and Software Program (WASP) funded by the Knut and Alice Wallenberg Foundation.

\subsection{Author contribution}
All authors contributed to the study conception and design. The software was developed by A.F and E.W. The methodology was developed by A.F. and T.L. The experiments were carried out by T.S. Supervision was done by T.L. and M.S. The first draft of the manuscript was written by A.F. Review and editing was done by A.F. and M.S. All authors read and approved the final manuscript.

\appendix
\section{Example output}
Examples of successful and failed output grasps when evaluated on test data. Targeted logs are shown in brighter colors. The shown grasp is the pile's top scoring grasp according to the model when assessed using the $f_1$ objective function.
\begin{figure*}[h]
    \centering
    \includegraphics[width=\textwidth]{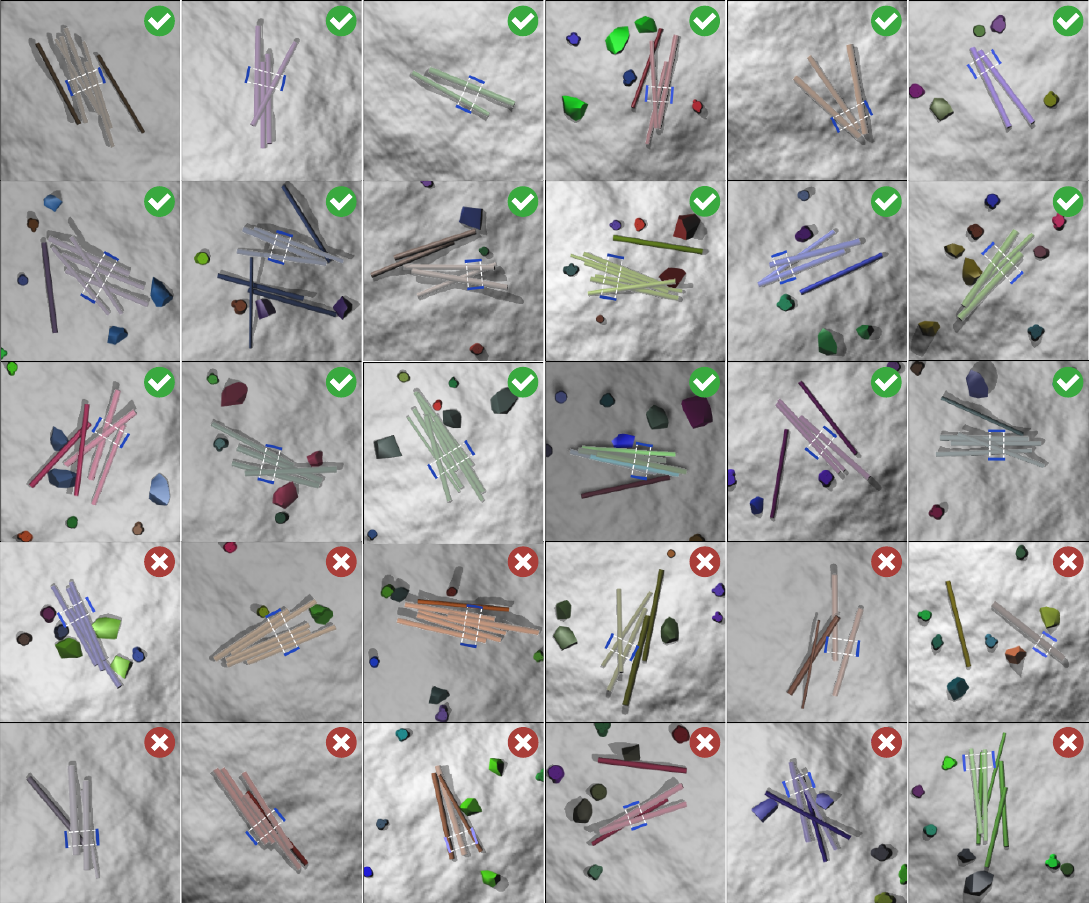}
    \caption{Examples of successful and failed model output when evaluated on test data. Fourth and fifth rows show grasps that failed due to collisions with surrounding objects, and grasping more logs than intended, respectively.}
\label{fig:success_and_fail}
\end{figure*}

\section{Depth ablation study}
The initial V0 model that was used in the experiments on the real platform only had access to RGB and target mask $M_T$ as input, no depth. Adding depth gave both a lower final loss during training and performed better in simulation. The test set simulation results are summarized in Table~\ref{tab:depth_ablation}. Note that the rows corresponding to the RGBD model are the same as the ones in Table~\ref{tab:results_2_to_7}.
\begin{table}[h]
    \centering
     \caption{Test set performance (success rate/average number of grasped logs) for the two considered models for different pile types. For each combination of pile size and obstacles (Yes/No), we simulate 100 piles, one grasp per pile, giving us 1,200 piles in total.}
    \begin{tabular}{m{1cm}m{1.3cm}m{1.4cm}m{1.1cm}m{1.1cm}m{1.1cm}m{1.1cm}m{1.1cm}}
    \toprule Model &
    Obstacles & 2 logs & 3 logs & 4 logs & 5 logs & 6 logs & 7 logs \\
    \midrule
    \multirow{2}{*}{RGBD} & Yes & 99\%/1.7 & 87\%/1.9 & 89\%/2.7 & 89\%/3.6 & 83\%/3.9 & 78\%/4.5 \\
    \cmidrule(lr){2-8}
    & No & 100\%/1.8 & 99\%/2.5 & 98\%/3.2 & 90\%/3.6 & 89\%/4.1 & 73\%/4.2 \\
    \midrule
    \multirow{2}{*}{RGB} & Yes & 97\%/1.5 & 96\%/2.0 & 86\%/2.5 & 82\%/3.1 & 79\%/3.2 & 69\%/3.2 \\
    \cmidrule(lr){2-8}
     & No & 100\%/1.7 & 98\%/2.2 & 93\%/2.8 & 84\%/3.0 & 91\%/3.8 & 70\%/3.5 \\
    \bottomrule
    \end{tabular}
    \label{tab:depth_ablation}
\end{table}

\end{document}